\documentclass[letterpaper, 10 pt, conference]{ieeeconf}  

\IEEEoverridecommandlockouts                              

\newcommand{\yenchen}[1]{\textcolor{black}{#1}}
\newcommand{\pete}[1]{{\color{black}{#1}}}
\newcommand{\ty}[1]{\textcolor{black}{#1}}

\usepackage{times}
\usepackage{epsfig}
\usepackage{graphicx}
\usepackage{amsmath}
\usepackage{amssymb}
\usepackage[hypcap=false]{caption}
\usepackage{subcaption}
\usepackage[dvipsnames]{xcolor}
\usepackage{paralist}
\usepackage[colorlinks=false]{hyperref}
\usepackage{commath}
\usepackage[font=small,labelfont=bf]{caption} 

\def\etal{et~al.~}			  
\def\eg{e.g.,~}               
\def\ie{i.e.,~}               

\newcommand{\numrays}{b}
\newcommand{\numpoints}{n}
\newcommand{\numimages}{N}

\title{\LARGE \bf
iNeRF: Inverting Neural Radiance Fields for Pose Estimation
}

\author{
  \begin{tabular}{ccc}
  Lin Yen-Chen$^{1,2}$ \qquad & Pete Florence$^1$ \qquad & Jonathan T. Barron$^1$ \\ Alberto Rodriguez$^2$ \qquad & Phillip Isola$^2$ \qquad & Tsung-Yi Lin$^1$ 
  \end{tabular}
  \vspace{0.1cm}
  \\
  $^1$Google Research \qquad $^2$Massachusetts Institute of Technology
  \\
}

\begin{document}
\twocolumn[{%
\renewcommand\twocolumn[1][]{#1}%
\maketitle
\vspace{-1cm}
\begin{center}
		{
		\vspace{0.0cm}
		\normalsize
}
\centering

\vspace{0.15cm}
\href{http://yenchenlin.me/inerf/assets/overview.gif}{
\includegraphics[width=\linewidth]{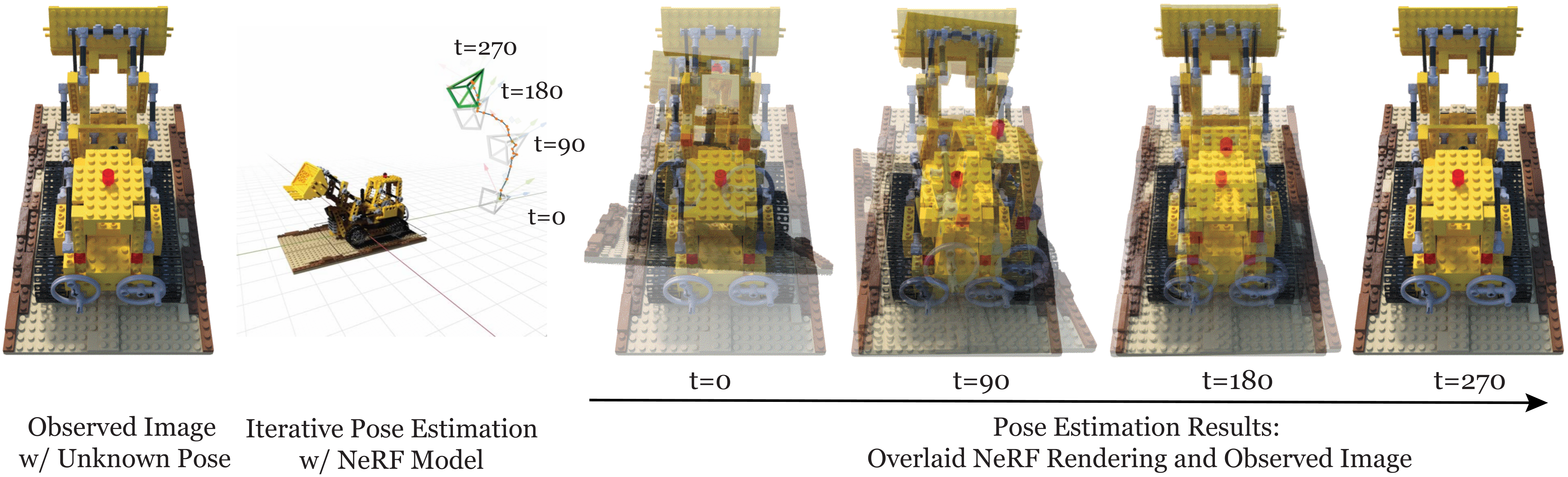}}
    \captionof{figure}{We present iNeRF which performs \yenchen{mesh-free} pose estimation by inverting a neural radiance field of an object or scene. The middle figure shows the trajectory of estimated poses (\textcolor{gray}{gray}) and the ground truth pose (\textcolor{LimeGreen}{green}) in iNeRF's iterative pose estimation procedure.  By comparing the observed and rendered images, we perform gradient-based optimization to estimate the camera's pose \yenchen{without accessing the object's mesh model}. \textbf{Click the image to play the video in a browser.}}    \label{fig:teaser}
\end{center}%
}]


\begin{abstract}

We present iNeRF, a framework that performs \yenchen{mesh-free} pose estimation by ``inverting'' a Neural Radiance Field (NeRF).
NeRFs have been shown to be remarkably effective for the task of view synthesis --- synthesizing photorealistic novel views of real-world scenes or objects. In this work, we investigate whether we can apply analysis-by-synthesis via NeRF for mesh-free, RGB-only 6DoF pose estimation -- given an image, find the translation and rotation of a camera relative to a 3D object or scene.
\yenchen{Our method assumes that no object mesh models are
available during either training or test time.}
Starting from an initial pose estimate, we use gradient descent to minimize the residual between pixels rendered from a 
NeRF and pixels in an observed image.
In our experiments, we first study 1) how to sample rays during pose refinement for iNeRF to collect informative gradients and 2) how different batch sizes of rays affect iNeRF on a synthetic dataset.
We then show that for complex real-world scenes from the LLFF dataset, iNeRF can improve NeRF by estimating the camera poses of novel images and using these images as additional training data for NeRF. 
\yenchen{Finally, we show iNeRF can perform category-level object pose estimation, including object instances not seen during training, with RGB images by inverting a NeRF model inferred from a single view.}

\end{abstract}

\section{INTRODUCTION}
Six degree of freedom (6DoF) pose estimation has a wide range of applications, including robot manipulation, and mobile robotics, and augmented reality, \cite{manuelli2019kpam, marion2018label, hodan2018bop}. Recent progress in differentiable rendering has sparked interest in solving pose estimation via analysis-by-synthesis~\cite{chen2020category,ma2020deep,park2020latentfusion,wang2020self6d}. However, techniques built around differentiable rendering engines typically require a high-quality watertight 3D model\yenchen{, \eg mesh model,} of the object for use in rendering. Obtaining such models can be difficult and labor-intensive, and objects with unusual transparencies, shapes, or material properties may not be amenable to 3D model formats used in rendering engines. 

\yenchen{The recent advances of} Neural Radiance Fields (NeRF~\cite{mildenhall2020nerf}) provide a mechanism for capturing complex 3D and optical structures from only one or a few RGB images, which opens up the opportunity to apply analysis-by-synthesis to broader real-world scenarios \yenchen{without mesh models during training or test times.} 
%
\ty{NeRF representations parameterize the density and color of the scene as a function of 3D scene coordinates. The function can either be learned from multi-view images with given camera poses~\cite{martin2020nerf,mildenhall2020nerf} or directly predicted by a generative model given one or few input images~\cite{wang2021ibrnet, yu2020pixelnerf}.}

Here we present iNeRF, a new framework for 6 DoF pose estimation by inverting a NeRF \yenchen{model}. . iNeRF takes three inputs: an observed image, an initial estimate of the pose, and a NeRF model representing a 3D scene or an object in the image. We adopt an analysis-by-synthesis approach to compute the appearance differences between the pixels rendered from the NeRF model and the pixels from the observed image. The gradients from these residuals are then backpropagated through the NeRF model to produce the gradients for the estimated pose. As illustrated in Figure~\ref{fig:teaser}, this procedure is repeated iteratively until the rendered and observed images are aligned, thereby yielding an accurate pose estimate.
Despite its compelling reconstruction ability, using NeRF as a differentiable renderer for pose estimation through gradient-based optimization presents several challenges.
For one, NeRF renders each pixel in an image by shooting a ray through that pixel and repeatedly querying a 3D radiance field (parameterized by a neural network) while marching along that ray  to produce estimates of volume density and color that are then alpha-composited into a pixel color.
This rendering procedure is expensive, which presents a problem for an analysis-by-synthesis approach which, naively, should require rendering a complete image and backpropagating the loss contributed by all pixels.
For iNeRF, we address this issue by capitalizing on the fact that NeRF's ray-marching structure allows rays and pixels to be rendered individually, and we present an interest point-based sampling approach that allows us to identify \emph{which rays} should be sampled to best inform the pose of the object. This sampling strategy allows for accurate pose estimation while using two orders of magnitude fewer pixels than a full-image sampling.
%
%
Furthermore, we demonstrate iNeRF can improve NeRF's reconstruction quality by annotating images without pose labels and adding them to the training set. We show that this procedure can reduce the number of required labeled images by 25\% while maintaining reconstruction quality.
%

\yenchen{Finally, we show iNeRF can perform category-level object pose estimation, including object instances not seen during training, with RGB inputs by inverting a NeRF model inferred by pixelNeRF~\cite{yu2020pixelnerf} given a single view of the object.}
\pete{The only prior work we are aware of that similarly provides RGB-only category-level pose estimation is the recent work of Chen et al. \cite{chen2020category}.  In Sec.~\ref{sec:related} we compare differences between \cite{chen2020category} and our work, which mostly arise from the opportunities and challenges presented by a continuous, implicit NeRF parameterization.}


To summarize, our primary contributions are as follows. (i) We show that iNeRF can use a 
NeRF model to estimate 6 DoF pose for scenes and objects with complex geometry, without the use of 3D mesh models or depth sensing --- only RGB images are used as input. (ii) We perform a thorough investigation of ray sampling and the batch sizes for gradient optimization to characterize the robustness and limitations of iNeRF. (iii) We show that iNeRF can improve NeRF by predicting the camera poses of additional images, that can then be added into NeRF's training set. \pete{(iv) We show category-level pose estimation results, for unseen objects, including a real-world demonstration.}

\section{RELATED WORKS}\label{sec:related}
\textbf{Neural 3D shape representations.} Recently, several works have investigated representing 3D shapes implicitly with neural networks. In this formulation, the geometric or appearance properties of a 3D point $\mathbf{x} = (x, y, z)$ is parameterized as the output of a neural network.
The advantage of this approach is that scenes with complex topologies can be represented at high resolution with low memory usage.
When ground truth 3D geometry is available as supervision, neural networks can be optimized to represent the signed distance function~\cite{park2019deepsdf} or occupancy function~\cite{mescheder2019occupancy}.
However, ground truth 3D shapes are hard to obtain in practice. 
This motivates subsequent work on relaxing this constraint by formulating differentiable rendering pipelines that allow neural 3D shape representations to be learned using only 2D images as supervision~\cite{kato2020differentiable,lin2020sdfsrn,lin2019photometric}.
Niemeyer~\etal~\cite{niemeyer2020differentiable} represent a surface as a neural 3D occupancy field and texture as a neural 3D texture field. Ray intersection locations are first computed with numerical methods using the occupancy field and then provided as inputs to the texture field to output the colors.
Scene Representation Networks~\cite{sitzmann2019scene} learn a neural 3D representation that outputs a feature vector and RGB color at each continuous 3D coordinate and employs a recurrent neural network to perform differentiable ray-marching.
NeRF~\cite{mildenhall2020nerf} shows that by taking view directions as additional inputs, a learned neural network works well in tandem with volume rendering techniques and enables photo-realistic view synthesis.
NeRF in the Wild~\cite{martin2020nerf} extends NeRF to additionally model each image's individual appearance and transient content, thereby allowing high-quality 3D reconstruction of landmarks using unconstrained photo collections.
NSVF~\cite{liu2020neural} improves NeRF by incorporating a sparse voxel octree structure into the scene representation, which accelerates rendering by allowing voxels without scene content to be omitted during rendering.
\yenchen{To generalize across scenes or objects, pixelNeRF~\cite{yu2020pixelnerf} and IBRNet~\cite{wang2021ibrnet} predict NeRF models conditioned on input images.}
Unlike NeRF and its variants, which learn to represent a scene's structure from posed RGB images, we address the inverse problem: how to localize new observations whose camera poses are unknown, using a 
NeRF.

\begin{figure*}[t!]
    \centering
    \includegraphics[scale=0.4]{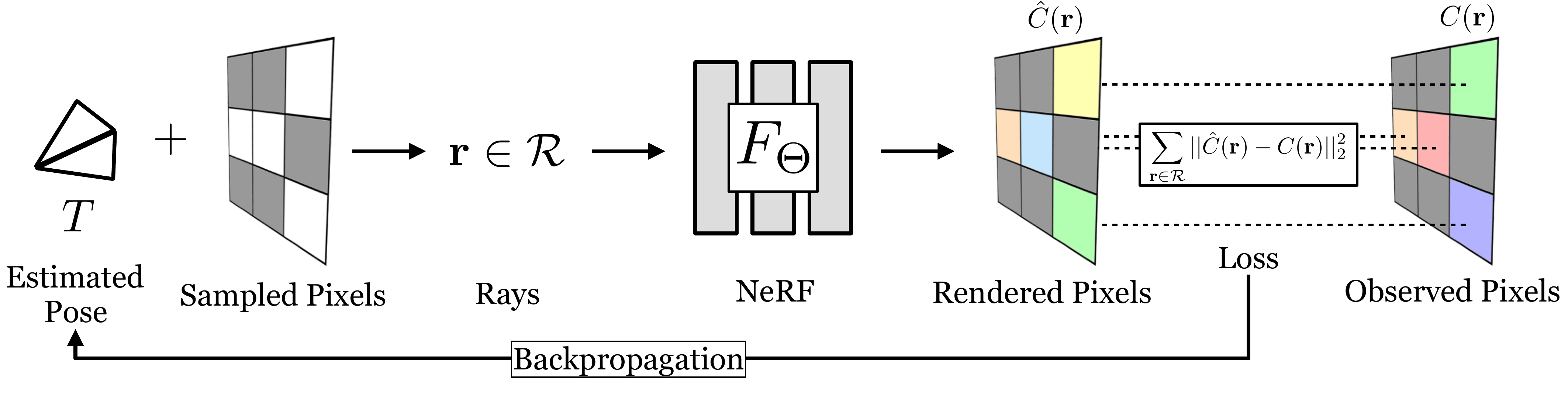}
    \caption{An overview of our pose estimation pipeline which inverts an optimized neural radiance field (NeRF). Given an initially estimated pose, we first decide which rays to emit. Sampled points along the ray and the corresponding viewing direction are fed into NeRF's volume rendering procedure to output rendered pixels. Since the whole pipeline is differentiable, we can refine our estimated pose by minimizing the residual between the rendered and observed pixels.}
    \label{fig:method}
\end{figure*}
%

\textbf{Pose Estimation from RGB Images.}
%
Classical methods for object pose estimation address the task by detecting and matching keypoints with known 3D models~\cite{aubry2014seeing,collet2011moped,ferrari2006simultaneous,rothganger20063d}. Recent approaches based on deep learning have proposed to 1) directly estimate objects pose using CNN-based architectures~\cite{schwarz2015rgb,tulsiani2015viewpoints,xiang2017posecnn} or 2) estimate 2D keypoints~\cite{pavlakos20176,suwajanakorn2018discovery,tekin2018real,tremblay2018deep} and solve for pose using the PnP-RANSAC algorithm. Differentiable mesh renderers~\cite{chen2019learning,palazzi2018end} have also been explored for pose estimation.
Although their results are impressive, all the aforementioned works require access to objects' 3D models during both training and testing, which significantly limits the applicability of these approaches.
\pete{Recently, Chen~\etal~\cite{chen2020category} address category-level object pose estimation~\cite{wang2019normalized}, in particular they impressively estimate object shape and pose across a category from a single image. They use a single-image reconstruction with a 3D \textit{voxel-based} feature volume and then estimating pose using iterative image alignment.
%
%
In contrast, in our work we use \textit{continuous} implicit 3D representations in the form of NeRF models, which have been empirically shown to produce more photorealistic novel-image rendering \cite{mildenhall2020nerf, martin2020nerf} and scale to large, building-scale volumes \cite{martin2020nerf}, which we hypothesize will enable higher-fidelity pose estimation.
This also presents challenges, however, due to the expensive computational cost of NeRF rendering, for which we introduce a novel importance-sampling approach in Sec.~\ref{subsec:sampling}.  Another practical difference in our approach to category-level pose estimation -- while \cite{chen2020category} optimizes for shape with gradient descent, we show we can instead allow pixelNeRF to predict a NeRF model with just a forward pass of a network.
Additionally, since NeRF models scale well to large scenes, we can use the same iNeRF formulation to perform localization, for example in challenging real-world LLFF scenes -- this capability was not demonstrated in \cite{chen2020category}, and may be challenging due to the memory limitations of voxel representations for sufficient fidelity in large scenes.}
While \textit{object pose estimation} methods are often separate from methods used for \textit{visual localization} of a camera in a scene as in the SfM literature (i.e. \cite{shotton2013scene,valentin2015exploiting,schmidt2016self}), because NeRF and iNeRF only require posed RGB images as training, iNeRF can be applied to localization as well. 

\section{Background}\label{background}
Given a collection of $\numimages$ RGB images $\{I_i\}_{i=1}^{\numimages}$, $I_i \in [0, 1]^{H \times W \times 3}$ with known camera poses $\{T_i\}_{i=1}^\numimages$, NeRF learns to synthesize novel views associated with unseen camera poses. 
NeRF does this by representing a scene as a ``radiance field'': a volumetric density that models the shape of the scene, and a view-dependent color that models the appearance of occupied regions of the scene, both of which lie within a bounded 3D volume.
The density $\sigma$ and RGB color $\mathbf{c}$ of each point are parameterized by the weights $\Theta$ of a multilayer perceptron (MLP) $F$ that takes as input the 3D position of that point $\mathbf{x} = (x, y, z)$ and the unit-norm viewing direction of that point $\mathbf{d} = (d_x, d_y, d_z)$, where $(\sigma, c) \leftarrow F_{\Theta}(\mathbf{x}, \mathbf{d})$.
To render a pixel, NeRF emits a camera ray from the center of the projection of a camera through that pixel on the image plane. Along the ray, a set of points are sampled for use as input to the MLP which outputs a set of densities and colors. These values are then used to approximate the image formation behind volume rendering~\cite{kajiya84} using numerical quadrature~\cite{max95}, producing an estimate of the color of that pixel. NeRF is trained to minimize a photometric loss $\mathcal{L} = \sum_{\mathbf{r} \in \mathcal{R}} ||\hat{C}(\mathbf{r}) - C(\mathbf{r})||_2^2$,
using some sampled set of rays $\mathbf{r} \in \mathcal{R}$ where $C(\mathbf{r})$ is the observed RGB value of the pixel corresponding to ray $\mathbf{r}$ in some image, and $\hat{C}(\mathbf{r})$ is the prediction produced from neural volume rendering. 
To improve rendering efficiency one may train two MLPs: one ``coarse'' and one ``fine'', where the coarse model serves to bias the samples that are used for the fine model.
For more details, we refer readers to Mildenhall \etal~\cite{mildenhall2020nerf}. 
\yenchen{Although NeRF originally needs to optimize the representation for every scene independently, several extensions~\cite{Rematas21arxiv_sharf,Trevithick20arxiv_GRF,wang2021ibrnet,yu2020pixelnerf} have been proposed to directly predict a continuous neural scene representation conditioned on one or few input images. 
In our experiments, we show that iNeRF can be used to perform 6D pose estimation with either an optimized or predicted NeRF model.}
%
%

\section{iNerf Formulation}
We now present iNeRF, a framework that performs 6 DoF pose estimation by ``inverting'' a trained NeRF. 
Let us assume that the NeRF of a scene or object parameterized by $\Theta$ has already been recovered and that the camera intrinsics are known, but the camera pose $T$ of an image observation $I$ are as-yet undetermined. 
Unlike NeRF, which optimizes $\Theta$ using a set of given camera poses and image observations, we instead solve the inverse problem of recovering the camera pose $T$ given the weights $\Theta$ and the image $I$ as input:
\begin{equation}~\label{eq:formulation}
\hat{T} = \underset{T \in \text{SE(3)}}{\text{argmin}} \ \mathcal{L}(T  \ | \ I, \Theta)
\end{equation}
To solve this optimization, we use the ability from NeRF to take some estimated camera pose $T \in$ SE(3) in the coordinate frame of the NeRF model and render a corresponding image observation. We can then use the same photometric loss function $\mathcal{L}$ as was used in NeRF (Sec.~\ref{background}), but rather than backpropagate to update the weights $\Theta$ of the MLP, we instead update the pose $T$ to minimize $\mathcal{L}$. The overall procedure is shown in Figure~\ref{fig:method}.
While the concept of inverting a NeRF to perform pose estimation can be concisely stated, it is not obvious that such a problem can be practically solved to a useful degree. The loss function $\mathcal{L}$ is non-convex over the 6DoF space of SE(3), and full-image NeRF renderings are computationally expensive, particularly if used in the loop of an optimization procedure.  Our formulation and experimentation (Sec.~\ref{sec:experiments}) aim to address these challenges. In the next sections, we discuss (i) the gradient-based SE(3) optimization procedure, (ii) ray sampling strategies, and (iii) how to use iNeRF's predicted poses to improve NeRF.



\subsection{Gradient-Based SE(3) Optimization}

Let $\Theta$ be the parameters of a trained and fixed NeRF, $\hat{T_i}$ the estimated camera pose at current optimization step $i$, $I$ the observed image, and $\mathcal{L}(\hat{T}_i \ | \ I, \Theta)$ be the loss used to train the fine model in NeRF.
We employ gradient-based optimization to solve for $\hat{T}$ as defined in Equation~\ref{eq:formulation}. 
To ensure that the estimated pose $\hat{T}_i$ continues to lie on the SE(3) manifold during gradient-based optimization, we parameterize $\hat{T}_i$ with exponential coordinates.
Given an initial pose estimate $\hat{T}_{0} \in$ SE(3) from the camera frame to the model frame, we represent $\hat{T}_{i}$ as:
\begin{gather*} 
\hat{T}_{i} = e^{[\mathcal{S}_i]\theta_i} \hat{T}_{0}\,, \\
\text{where}\quad e^{[\mathcal{S}]\theta} = 
\begin{bmatrix} e^{[\omega]\theta} & K(\mathcal{S}, \theta) \\ 0 & 1 \end{bmatrix}\,,
\end{gather*}
where $\mathcal{S} = [\omega,  \nu]^{\mathrm{T}}$ represents the screw axis, $\theta$ the magnitude, $[w]$ represents the skew-symmetric $3 \times 3$ matrix of $w$, and $K(\mathcal{S}, \theta) = (I\theta + (1-\cos \theta)[\omega] + (\theta - \sin \theta)[\omega]^2)\nu$~\cite{lynch2017modern}.
With this parameterization, our goal is to solve the optimal relative transformation from an initial estimated pose $T_0$:
\begin{equation}~\label{eq:exp_formulation}
\widehat{\mathcal{S}\theta} = \underset{S\theta \in \mathbb{R}^6}{\text{argmin}} \ \mathcal{L}(e^{[\mathcal{S}]\theta} T_{0}   \ | \ I, \Theta).
\end{equation}
We iteratively differentiate the loss function through the MLP to obtain the gradient $\nabla_{\mathcal{S}\theta}\mathcal{L}(e^{[\mathcal{S}]\theta} T_{0}   \ | \ I, \Theta)$ 
that is used to update the estimated relative transformation.  We use Adam optimizer \cite{kingma2014adam} with an exponentially decaying learning rate (See Supplementary for parameters).
For each observed image, we initialize $\mathcal{S}\theta$ near $\mathbf{0}$, where each element is drawn at random from a zero-mean normal distribution $\mathcal{N}(0, \sigma=10^{-6})$.
%
%
%
In practice, parameterizing with $e^{[\mathcal{S}]\theta} \ T_{\text{0}}$ rather than $T_{\text{0}} \ e^{[\mathcal{S}]\theta}$ results in a center-of-rotation at the initial estimate's center, rather than at the camera frame's center. This alleviates coupling between rotations and translations during optimization.

\subsection{Sampling Rays}\label{subsec:sampling}
In a typical differentiable render-and-compare pipeline, one would want to leverage the gradients contributed by all of the output pixels in the rendered image~\cite{wang2020self6d}.
However, with NeRF, each output pixel's value is computed by weighing the values of $\numpoints$ sampled points along each ray $\mathbf{r} \in \mathcal{R}$ during ray marching, so given the amount of sampled rays in a batch $\numrays = |\mathcal{R}|$, then $\mathcal{O}(\numrays \numpoints)$ forward/backward passes of the underlying NeRF MLP will be queried.
%
%
Computing and backpropagating the loss of all pixels in an image (\ie, $\numrays=HW$, where $H$ and $W$ represent the height and width of a high-resolution image) therefore require significantly more memory than is present on any commercial GPU.  While we may perform multiple forward and backward passes to accumulate these gradients, this becomes prohibitively slow to perform each step of our already-iterative optimization procedure. In the following, we explore strategies for selecting a sampled set of rays $\mathcal{R}$ for use in evaluating the loss function $\mathcal{L}$ at each optimization step.  In our experiments we find that we are able to recover accurate poses while  sampling only $\numrays=2048$ rays per gradient step, which corresponds to a single forward/backward pass that fits within GPU memory and provides $150\times$ faster gradient steps on a $640 \times 480$ image.

\begin{figure}
    \centering
    \includegraphics[width=\linewidth]{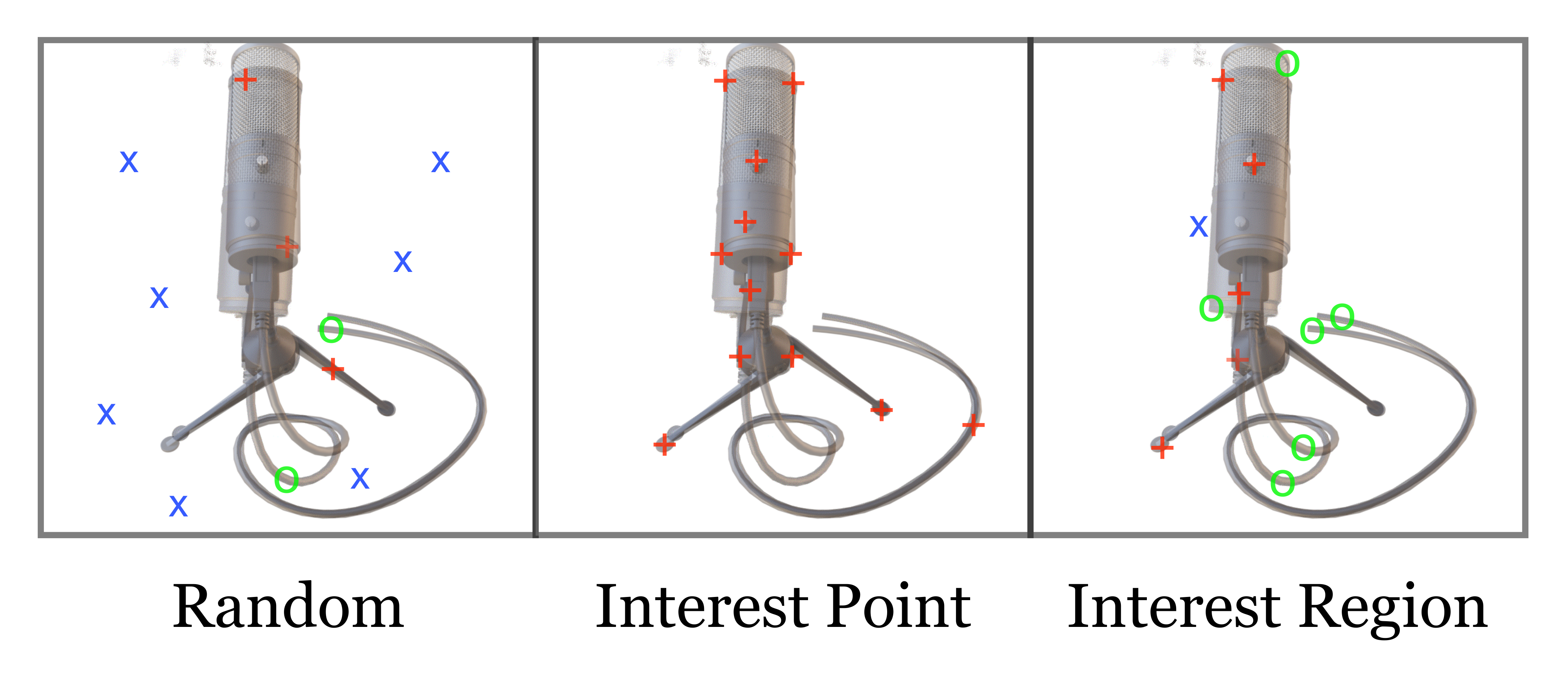}
    \caption{An illustration of 3 sampling strategies. The input image and the rendering corresponding to the estimated pose of the scene are averaged. We use \textcolor{blue}{x} to represent sampled pixels on the background; \textcolor{red}{+} to represent sampled pixels that are covered by both rendered and observed images; \textcolor{green}{o} to represent sampled pixels that are only covered by either the rendered or the input image. When performing random sampling (left) many sampled pixels are \textcolor{blue}{x}, which provide no gradients for updating the pose. For ``interest point'' sampling (middle) some of the sampled pixels are already aligned and therefore provide little information. For ``interest region'' sampling, many sampled pixels are \textcolor{green}{o}, which helps pose estimation achieve higher accuracy and faster convergence.}
    \label{fig:sampling}
\end{figure}

\paragraph{Random Sampling.}
An intuitive strategy is to sample $M$ pixel locations $\{p^i_x, p^i_y\}_{i=0}^{M}$ on the image plane randomly and compute their corresponding rays. Indeed,  NeRF itself uses this strategy when optimizing $\Theta$ (assuming image batching is not used).
We found this random sampling strategy's performance to be ineffective when the batch size of rays $\numrays$ is small. Most randomly-sampled pixels correspond to flat, textureless regions of the image, which provide little information with regards to pose (which is consistent with the well-known aperture problem~\cite{wallach1935visuell}). See Figure~\ref{fig:sampling} for an illustration. 
\paragraph{Interest Point Sampling.}
Inspired by the literature of image alignment~\cite{szeliski2006image}, we propose interest point sampling to guide iNeRF optimization, where we first employ interest point detectors to localize a set of candidate pixel locations in the observed image.
We then sample $M$ points from the detected interest points and fall back to random sampling if not enough interest points are detected.
Although this strategy makes optimization converge faster since less stochasticity is introduced, we found that it is prone to local minima as it only considers interest points on the observed image instead of interest points from both the observed and rendered images.
However, obtaining the interest points in the rendered image requires $\mathcal{O}(HWn)$ forward MLP passes and thus prohibitively expensive to be used in the optimization.

\paragraph{Interest Region Sampling.}
To prevent the local minima caused by only sampling from interest points, we propose using ``Interest Region'' Sampling, a strategy that relaxes Interest Point Sampling and samples from the dilated masks centered on the interest points.
After the interest point detector localizes the interest points, we apply a $5 \times 5$ morphological dilation for $I$ iterations to enlarge the sampled region.
In practice, we find this to speed up the optimization when the batch size of rays is small.
Note that if $I$ is set to a large number, Interest Region Sampling falls back to Random Sampling.
\subsection{Self-Supervising NeRF with iNeRF}

In addition to using iNeRF to perform pose estimation given a trained NeRF, we also explore using the estimated poses to feed back into training the NeRF representation.  Specifically, we first (1) train a NeRF given a set of training RGB images with known camera poses $\{(I_i, T_i)\}_{i=1}^{N_{\text{train}}}$, yielding NeRF parameters $\Theta_{\text{train}}$. We then (2) use iNeRF to take in additional unknown-pose observed images $\{I_i\}_{i=1}^{N_{\text{test}}}$ and solve for estimated poses $\{\hat{T}_i\}_{i=1}^{N_{\text{test}}}$.  Given these estimated poses, we can then (3) use the self-supervised pose labels to add $\{(I_i, \hat{T}_i)\}_{i=1}^{N_{\text{test}}}$ into the training set.  This procedure allows NeRF to be trained in a semi-supervised setting.

\section{Results}\label{sec:experiments}
\begin{figure*}[t!]
    \centering
    \href{http://yenchenlin.me/inerf/assets/sampling.gif}{
    \includegraphics[scale=0.2]{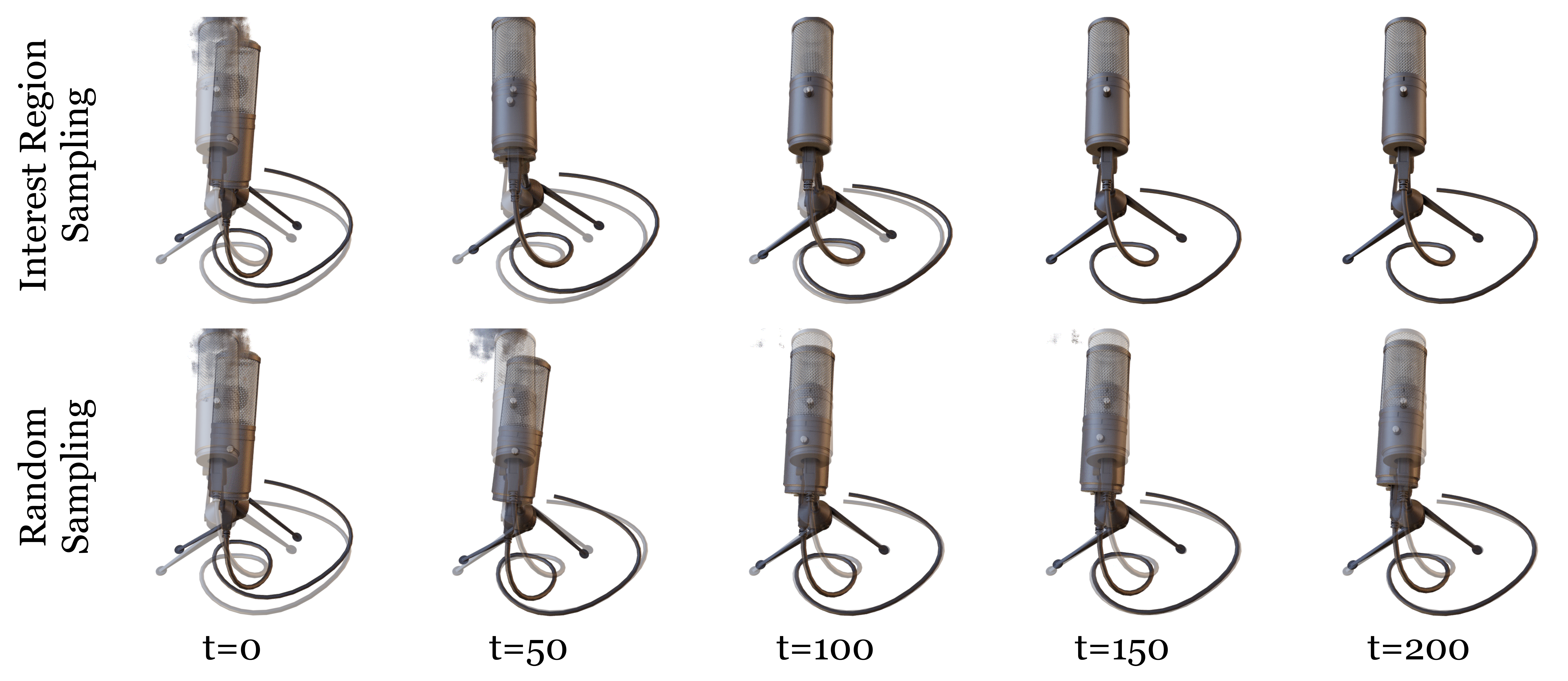}}
    \caption{We visualize the average of rendered images based on the estimated pose at time $t$ and the test image to compare different sampling methods.
    Adopting Interest Region Sampling helps our method to recover camera poses that align the rendered and test image to fine details. Random Sampling aligns the mic's rigging, but fails to align the wire. \textbf{Click the image to play the video in a browser.}}
    \label{fig:qual_blender}
\end{figure*}
%

We first conduct extensive experiments on the synthetic dataset from NeRF~\cite{mildenhall2020nerf} and the real-world complex scenes from LLFF~\cite{mildenhall2019local} to evaluate iNeRF for 6DoF pose estimation.
Specifically, we study how the batch size of rays and sampling strategy affect iNeRF.
We then show that iNeRF can improve NeRF by estimating the camera poses of images with unknown poses and using these images as additional training data for NeRF.
%
%
\yenchen{Finally, we show that iNeRF works well in tandem with pixelNeRF~\cite{yu2020pixelnerf} which predicts a NeRF model conditioned on a single RGB image. We test our method for category-level object pose estimation in both simulation and the real world. We found that iNeRF achieving competitive results against feature-based methods without accessing object mesh models during either training or test time.}

\subsection{Synthetic Dataset}~\label{exp:synthetic}
\vspace{-0.2cm}
\paragraph{Setting}
We test iNeRF on 8 scenes from NeRF's synthetic dataset. For each scene, we choose 5 test images and generate 5 different pose initializations by first randomly sampling an axis from the unit sphere and rotating the camera pose around the axis by a random amount within $[-40, 40]$ degrees. Then, we translate the camera along each axis by a random offset within $[-0.2, 0.2]$ meters.
\paragraph{Results}
We report the percentage of predicted poses whose error is less than 5$^{\circ}$ or 5cm at different numbers of steps. It is a metric widely used in the pose estimation community~\cite{hodan2018bop}.
Quantitative results are shown in 
Figure~\ref{fig:quan_blender}. We verify that under the same sampling strategy, larger batch sizes of rays achieve not only better pose estimation accuracy, but also faster convergence.
On the other hand, when the batch size of rays is fixed, interest region sampling is able to provide better accuracy and efficiency.
Specifically, the qualitative results shown in Figure~\ref{fig:qual_blender} clearly indicate that random sampling is inefficient as many sampled points lie on the common background and therefore provide no gradient for matching.
%

%
\subsection{LLFF Dataset}~\label{exp:llff}
\begin{figure*}
    \centering
    \begin{subfigure}[b]{0.9\textwidth}
    \includegraphics[width=1.0\textwidth]{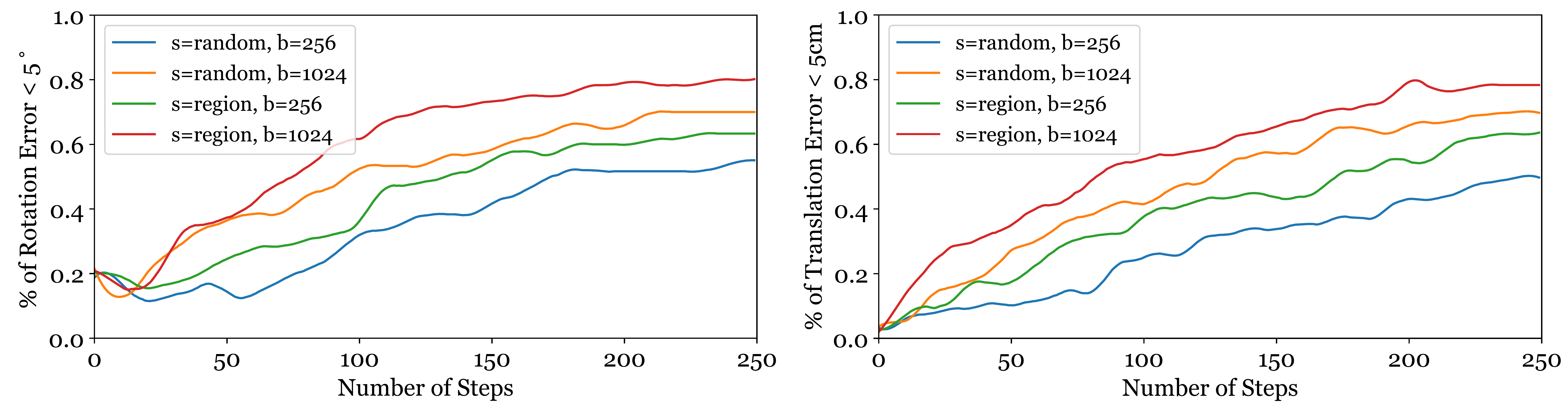}
    \caption{Results on the synthetic dataset.}
    \label{fig:quan_blender}
    \end{subfigure}
    \begin{subfigure}[b]{0.9\textwidth}
    \includegraphics[width=1.0\textwidth]{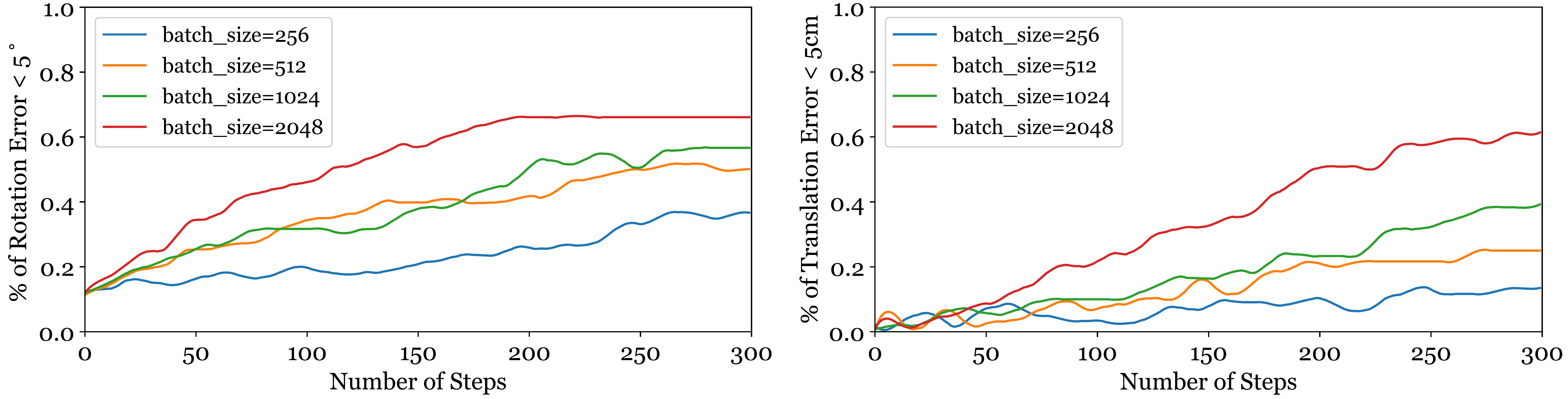}
    \caption{Results on real-world scenes from the LLFF dataset.}
    \label{fig:quan_LLFF}
    \end{subfigure}
    \caption{(a) Quantitative results on the synthetic dataset.  ``s'' stands for sampling strategy and ``b'' stands for the batch size. Applying Interest Region Sampling (s=region) improves the accuracy by 15\% across various batch sizes. (b) Quantitative results on LLFF.  Interest Region Sampling is always applied and we show the effect of various batch sizes on performance. Larger batch sizes can improve accuracy while reducing the number of gradient steps needed for convergence.}
    \label{fig:quan_blender_and_LLFF}
\end{figure*}

\vspace{-0.2cm}
\paragraph{Setting}
We use 4 complex scenes: \textit{Fern}, \textit{Fortress}, \textit{Horns}, and \textit{Room} from the LLFF dataset~\cite{mildenhall2019local}. For each test image, we generate 5 different pose initializations following the procedures outlined in Section~\ref{exp:synthetic} but instead translate the camera along each axis by a random offset within $[-0.1, 0.1]$ meters. Unlike the synthetic dataset where the images are captured on a surrounding hemisphere, images in the LLFF dataset are all captured with a forward-facing handheld cellphone.
%
\paragraph{Pose Estimation Results}
The percentage of predicted poses whose error is less than 5$^{\circ}$ or 5cm at different number of steps is reported in Figure~\ref{fig:quan_LLFF}. 
Similar to Section~\ref{exp:synthetic}, we find that the batch size of rays significantly affects iNeRF's visual localization performance.
Also, we notice that iNeRF performs worse on the LLFF dataset compared to the synthetic dataset. When the batch size of rays is set to 1024, the percentage of $< 5^{\circ}$  rotation errors drops from 71\% to 55\%, and the percentage of $< 5cm$ translation errors drops from 73\% to 39\%.
This difference across datasets may be due to the fact that the LLFF use-case in NeRF uses a normalized device coordinate (NDC) space, or may simply be a byproduct of the difference in scene content.
%

\begin{figure*}
    \centering
    \includegraphics[scale=0.28]{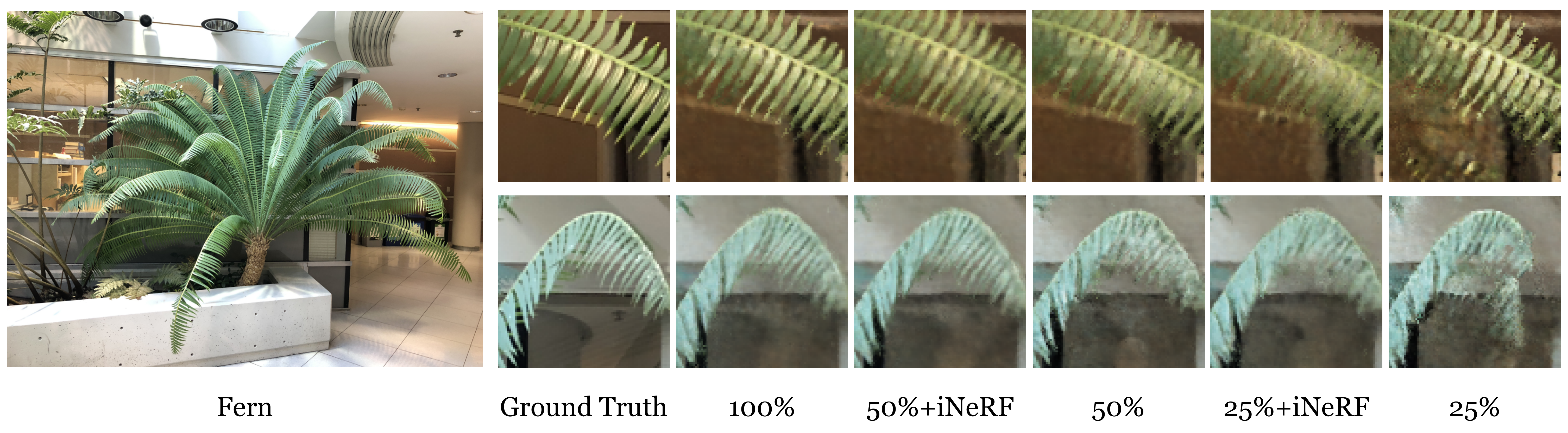}
    \caption{iNeRF can be used to improve NeRF by augmenting training data with images whose camera poses are unknown. We present an ablation study using 25\% and 50\% of training images to train NeRF models. These models are compared with models trained using 100\% of the training images but where a fraction of that data use estimated poses from iNeRF rather than ground-truth poses from the dataset.}
    \label{fig:ssl_nerf}
\end{figure*}

\paragraph{Self-Supervising NeRF with iNeRF Results}
We take the \textit{Fern} scene from the LLFF dataset and train it with 25\%, 50\%, and 100\% of the training data, respectively.
Then, NeRFs trained with 25\% and 50\% data are used by iNeRF to estimate the remaining training images' camera poses.
The estimated camera poses, together with existing camera poses, are used as supervision to re-train NeRF from scratch.
We report PSNRs in Table~\ref{table:ssl_nerf}. All of the models are trained for 200k iterations using the same learning rate.
We find that models that use the additional data made available through the use of iNeRF's estimated poses perform better. This finding is consistent with NeRF's well-understood sensitivity to the pose of its input cameras being accurate~\cite{mildenhall2020nerf}. Qualitative results can be found in Figure~\ref{fig:ssl_nerf}.

\begin{table}[]
\centering
\begin{tabular}{|l|l|cl|cl}
\hline
Label Fraction    & 100\%                   & \multicolumn{2}{c|}{50\%}                            & \multicolumn{2}{c|}{25\%}                           \\ \hline
iNeRF Supervision & \multicolumn{1}{c|}{No} & Yes                        & \multicolumn{1}{c|}{No} & Yes                    & \multicolumn{1}{c|}{No}    \\ \hline
PSNR              & 24.94                   & \multicolumn{1}{l|}{\textbf{24.64}} & 24.18                   & \multicolumn{1}{l|}{\textbf{23.89}} & \multicolumn{1}{l|}{21.85} \\ \hline
\end{tabular}

\caption{Benchmark on \textit{Fern} scene. NeRFs trained with pose labels generated by iNeRF can achieve higher PSNR.} 
\label{table:ssl_nerf}
\vspace{-0.5cm}
\end{table}
%


\subsection{ShapeNet-SRN Cars}~\label{exp:shapenet}
\vspace{-0.2cm}
\paragraph{Setting}
We evaluate iNeRF for RGB-based category-level object pose estimation.
Given two images $I_0$ and $I_1$ of an unseen instance, we seek to estimate its relative pose $T_0^1 = (R, t) \in SO(3) \times \mathbb{S}^2$ between images. The translation $t \in \mathbb{S}^2 \dot{=} \{t \in \mathbb{R}^3 \mid \norm{t} = 1\}$ is ambiguous up to a scale.
We perform the experiments on the “car” classes of ShapeNet using the dataset introduced in Stizmann~\etal\cite{sitzmann2019scene}.
The dataset contains 3514 cars that are split for training, validation, and test set. All the images have resolution of $128\times128$. The test images of the objects are rendered from 251 views on an archimedean spiral. For each object in the test set, $I_0$ is selected randomly from one of the 251 views and the other image $I_1$ is selected from views whose rotation and translation are within 30-degree from $I_0$. At test time, our method uses a pre-trained pixelNeRF to predict a NeRF model conditioned on image $I_0$.
Then, we apply iNeRF to align against $I_1$ for estimating the relative pose $T_0^1$.

\paragraph{Pose Estimation Results}
As shown in Table~\ref{table:shapenet}, our method achieves lower rotation and translation errors than a strong feature-based baseline, using SuperGlue \cite{sarlin2020superglue}. Importantly, iNeRF receives much fewer outliers: 8.7\% vs. 33.3\%. (Outliers are defined in Table~\ref{table:shapenet}). We note that in our method, our pose estimate is defined relative to a reference view of the object -- this is in contrast to \cite{chen2020category}, which depends on a canonical pose definition -- the subtleties of canonical pose definitions are discussed in \cite{manuelli2019kpam, wang2019normalized}. While \cite{chen2020category}'s method could be used in our setting, it would not make use of the reference image, and code was not available to run the comparison.   

\begin{table}[t]
\begin{center}
\begin{tabular}{c|cc|cc|c}
 & \multicolumn{2}{c|}{Rotation ($^\circ$)} & \multicolumn{2}{c|}{Translation ($^\circ$)} & Outlier
\\
Methods & Mean & Median & Mean & Median  & \%
\\ \hline
SuperGlue~\cite{sarlin2020superglue}      & 9.27     & 6.22     & 18.2     & 5.4 & 33.3
\\ \hline
Ours       & 4.39     & 2.01     &   4.81  & 1.77 & 8.7
\\ \hline
\end{tabular}
\caption{Quantitative results for the ShapeNet Cars dataset. We report performance using the mean and median of the translation and rotation error. A prediction is defined as an outlier when either the translation error or the rotation error is larger than 20$^\circ$.}
\vspace{-0.5cm}
\label{table:shapenet}
\end{center}
\end{table}

\subsection{Sim2Real Cars}~\label{exp:sim2real}
\begin{figure}[t!]
    \centering
    \includegraphics[scale=0.3]{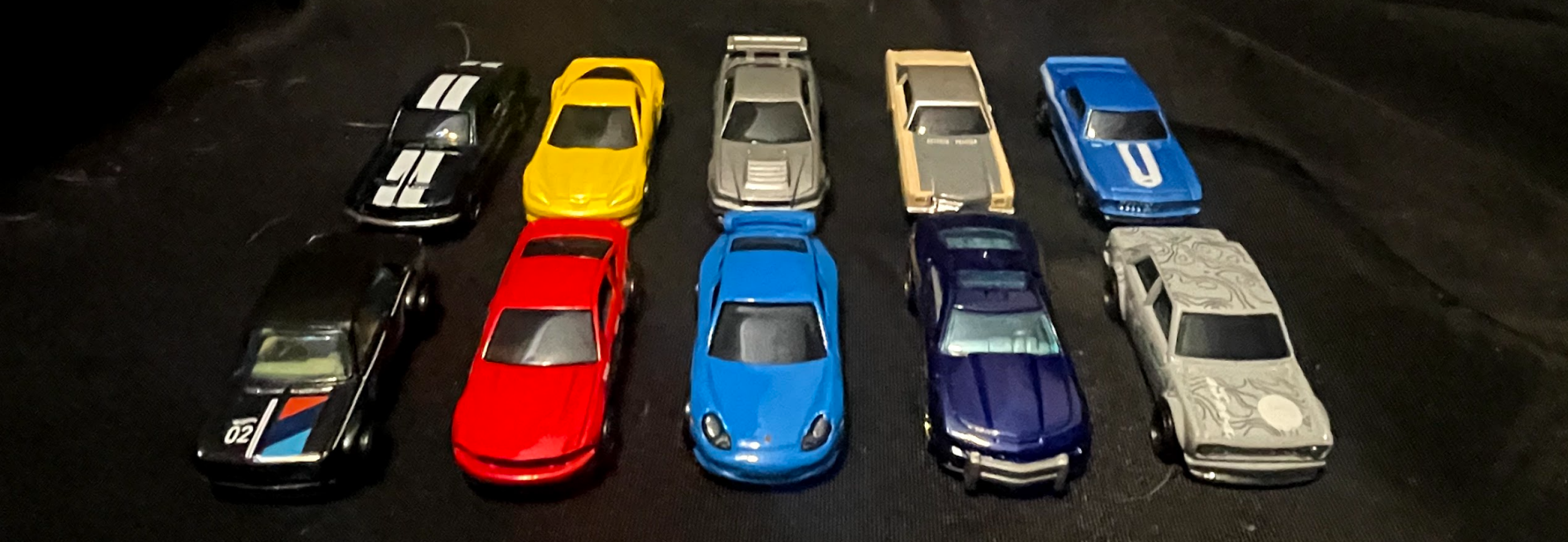}
    \caption{Real-world test data, toy cars with unknown mesh models.}
    \label{fig:real_cars}
\end{figure}

\begin{figure}[t!]
    \centering
    \href{http://yenchenlin.me/inerf/assets/object-level-slam.gif}{
    \includegraphics[scale=0.15]{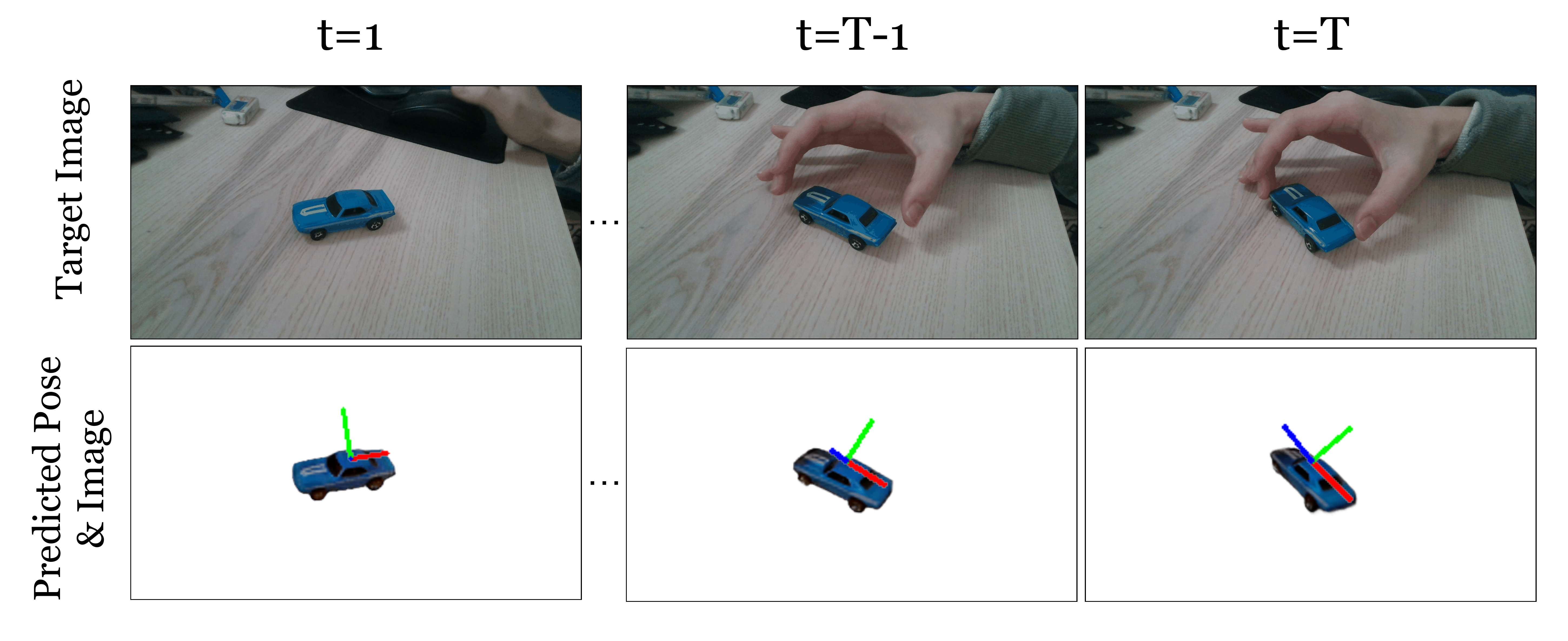}}
    \caption{Qualitative results of pose tracking in real-world images without the need for mesh/CAD model. In the left column, we show input video frames at different time steps. At each time $t$, iNeRF leverages a NeRF model inferred by pixelNeRF based on input frame at time $t-1$ to estimate the object's pose. In the right column, we show the resulting reconstructed frames and the estimated poses at each time step. The background has been masked out using PointRend~\cite{kirillov2020pointrend} before feeding the frame into pixelNeRF. The views are rotations about the view-space vertical axis.\textbf{Click the image to play the video in a browser.}}
    \label{fig:pose_tracking}
\end{figure}

\vspace{-0.2cm}
\paragraph{Setting}
We explore the performance of using iNeRF to perform category-level object pose estimation on real-world images. 10 unseen cars, as shown in Figure~\ref{fig:real_cars} are used as the test data.
\ty{We apply a pixelNeRF model trained on synthetic ShapeNet dataset to infer the NeRF model from a single real image without extra fine-tuning. Then we apply iNeRF with the inferred NeRF model to estimate the relative pose to the target real image.}
\paragraph{Pose Estimation Results}
We show the qualitative results of pose tracking in Figure~\ref{fig:pose_tracking}. At  each  time step $t$, iNeRF inverts a NeRF model conditioned on the frame at time $t-1$ to estimate the object’s pose. The reconstructed frame and estimated poses are also visualized. Since pixelNeRF requires a segmented image as input, we use PointRend~\cite{kirillov2020pointrend} to remove the background for frames that pixelNeRF takes as inputs.  In this iterative tracking setting, iNeRF only requires less than 10 iterations of optimization to converge which enables tracking at approximately 1Hz.

\section{Limitations and Future Work}
While iNeRF has shown promising results on pose estimation, it is not without limitations.
Both lighting and occlusion can severely affect the performance of iNeRF and are not modeled by our current formulation. One potential solution is to model appearance variation using transient latent codes as was done in NeRF-W~\cite{martin2020nerf} when training NeRFs, and jointly optimize these appearance codes alongside camera pose within iNeRF. 
%
%
Also, currently iNeRF takes around 20 seconds to run 100 optimization steps, which prevents it from being practical for real-time use. We expect that this issue may be mitigated with recent improvements in NeRF's rendering speed~\cite{liu2020neural}.
\section{Conclusion}
We have presented iNeRF, a framework for mesh-free, RGB-only pose estimation that works by inverting a NeRF model. We have demonstrated that iNeRF is able to perform accurate pose estimation using gradient-based optimization. We have thoroughly investigated how to best construct mini-batches of sampled rays for iNeRF and have demonstrated its performance on both synthetic and real datasets. Lastly, we have shown how iNeRF can perform category-level object pose estimation and track pose for novel object instances with an image conditioned generative NeRF model.


\section*{Appendix}
%
\section{Implementation Details}
\paragraph{Adam Optimizer and Learning Rate Schedule.}
The hyperparameter $\beta_1$ is set to 0.9 and $\beta_2$ is set to 0.999 in Adam optimizer.
%
We set the initial learning rate $\alpha_0$ to $0.01$. The learning rate at step $t$ is set as follow:
\begin{equation*}
    \alpha_t = \alpha_0 0.8 ^ {t / 100}\,. 
\end{equation*}
\paragraph{Loss for LineMOD.}
To compare the rendered and observed images, we transform both of them from RGB to YUV space using the standard definition:
\begin{equation*}
\begin{bmatrix}
Y \\
U \\
V \\
\end{bmatrix} = 
\begin{bmatrix}
0.2126 & 0.7152 & 0.0722\\
-0.09991 & -0.33609 & 0.436\\
0/615 & -0.55861 & -0.05639\\
\end{bmatrix}
\begin{bmatrix}
R \\
G \\
B \\
\end{bmatrix}
\end{equation*}
The Y channel is not considered in the computation of loss.
\section{Histogram of Pose Errors}
\begin{figure}[t!]
    \centering
    \includegraphics[width=0.45\textwidth]{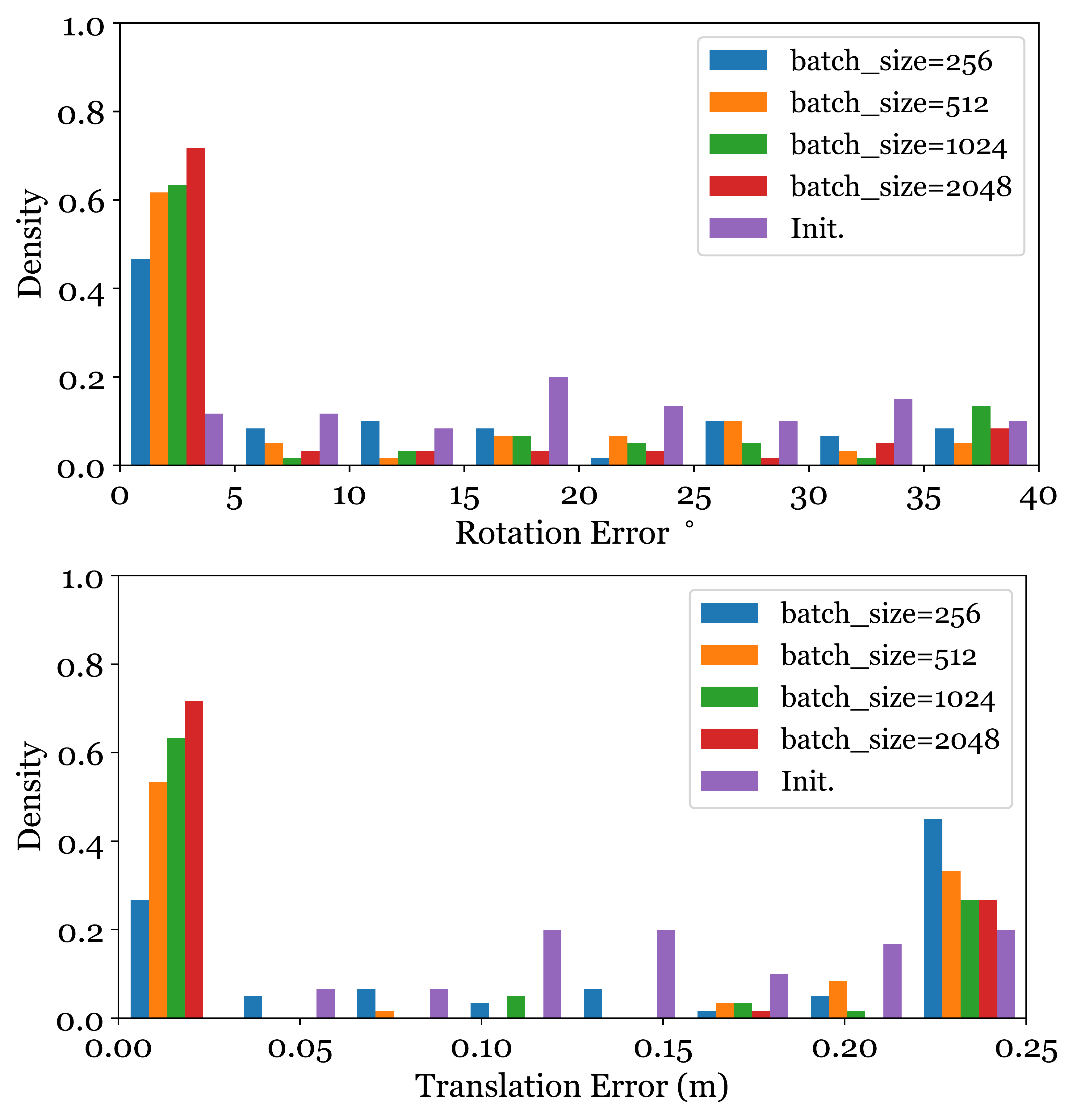}
    \caption{Histogram of pose errors on real-world scenes from the LLFF dataset.}
    \label{fig:hist_LLFF}
    \vspace{-0.4cm}
\end{figure}

We visualize the histogram of pose errors, before and after iNeRF optimization, on the LLFF dataset in Figure~\ref{fig:hist_LLFF} using the data from Section 5.2.
The data is generated by applying random perturbations within $[-40, 40]$ degrees for rotation and $[-0.1, 0.1]$ meters along each axis for translation. 
Note that when the batch size is 2048, more than 70\% of the data has $< 5^{\circ}$ and $< 5$ cm error after iNeRF is applied.

\section{More analysis in self-supervised NeRF}
For the \textit{Fern} scene, we found that when only 10\% of labeled camera poses are used, it worsens the PSNR from $18.5$ to $15.64$. The results show that having enough labels for a good initalization is important.
%








\bibliography{egbib}
\bibliographystyle{ieee_fullname}



\end{document}